\documentclass[10pt,twocolumn,letterpaper]{article}

\usepackage{cvpr}
\usepackage{times}
\usepackage{epsfig}
\usepackage{graphicx}
\usepackage{amsmath}
\usepackage{amssymb}
\usepackage{color}

\usepackage{authblk}
\makeatletter
\renewcommand\AB@affilsepx{; \protect\Affilfont}

\usepackage{caption}
\usepackage{subcaption}

\usepackage{tabulary}
\newcolumntype{K}[1]{>{\centering\arraybackslash}p{#1}}

\usepackage[breaklinks=true,bookmarks=false]{hyperref}

\cvprfinalcopy 

\def\cvprPaperID{747} 

\ifcvprfinal\fi
\begin{document}


\title{MAN: Moment Alignment Network for Natural Language Moment Retrieval via Iterative Graph Adjustment}
\author[$\dag$]{Da Zhang}
\author[$\ddag$]{Xiyang Dai}
\author[$\dag$]{Xin Wang}
\author[$\dag$]{Yuan-Fang Wang}
\author[$\S$]{Larry S. Davis}

\affil[$\dag$]{University of California, Santa Barbara}
\affil[$\ddag$]{Microsoft}
\renewcommand\AB@affilsepx{\\}
\affil[$\S$]{University of Maryland, College Park}

\affil[ ]{\tt\small \{dazhang, xwang, yfwang\}@cs.ucsb.edu, xiyang.dai@microsoft.com, lsd@umiacs.umd.edu}

\maketitle
\thispagestyle{empty}

\begin{abstract}
    This research strives for natural language moment retrieval in long, untrimmed video streams. The problem is not trivial especially when a video contains multiple moments of interests and the language describes complex temporal dependencies, which often happens in real scenarios. We identify two crucial challenges: semantic misalignment and structural misalignment. However, existing approaches treat different moments separately and do not explicitly model complex moment-wise temporal relations. In this paper, we present Moment Alignment Network (MAN), a novel framework that unifies the candidate moment encoding and temporal structural reasoning in a single-shot feed-forward network. MAN naturally assigns candidate moment representations aligned with language semantics over different temporal locations and scales. Most importantly, we propose to explicitly model moment-wise temporal relations as a structured graph and devise an iterative graph adjustment network to jointly learn the best structure in an end-to-end manner. We evaluate the proposed approach on two challenging public benchmarks DiDeMo and Charades-STA, where our MAN significantly outperforms the state-of-the-art by a large margin.
\end{abstract}

\begin{figure}
\begin{center}
\includegraphics[width=0.95\linewidth]{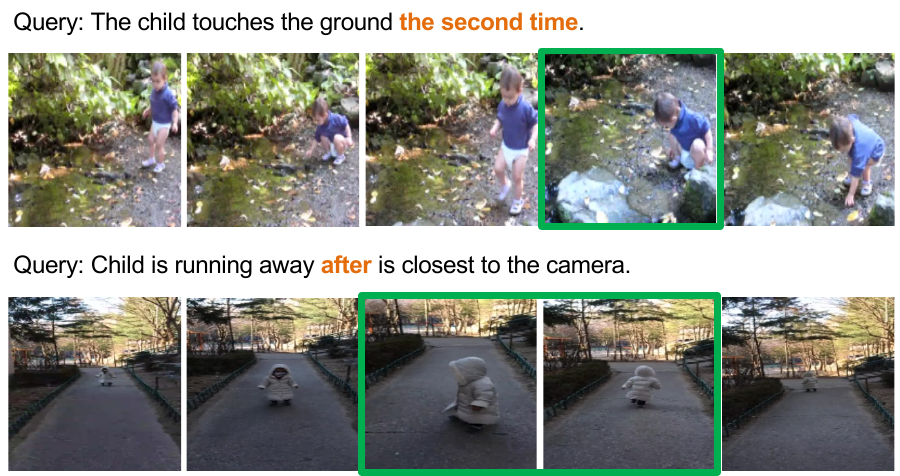}
\end{center}
   \caption{We consider the natural language moment retrieval task in untrimmed videos. To properly localize the moment, the retrieval model must handle both \textit{semantic misalignment (top)} with multiple moments of interests and \textit{structural misalignment (bottom)} with complex temporal dependencies.}
\label{fig:intro}
\vspace{-0.1cm}
\end{figure}

\section{Introduction}

    Video understanding is a fundamental problem in computer vision and has drawn increasing interests over the past few years due to its vast potential applications in surveillance, robotics, etc. While fruitful progress~\cite{simonyan2014two,wang2015towards,tran2015learning,carreira2017quo, tran2018closer, NonLocal2018,chao2018rethinking,wang2019learning,wang2018video, lin2017single,dai2017temporal,shou2017cdc,buch2017end,zhao2017temporal,shou2016temporal,xu2017r,zhang2018s3d} has been made on activity detection to recognize and localize temporal segments in videos, such approaches are limited to work on pre-defined lists of simple activities, such as playing basketball, drinking water, etc. This restrains us from moving towards real-world unconstrained activity detection. To solve this problem, we tackle the natural language moment retrieval task. Given a verbal description, our goal is to determine the start and end time (\ie localization) of the temporal segment (\ie moment) that best corresponds to this given query. While this formulation opens up great opportunities for  better video perception, it is substantially more challenging as it needs to model not only the characteristics of sentence and video but also their complex relations.
    
    On one hand, a real-world video often contains multiple moments of interests. Consider a simple query like ``The child touches the ground the second time", shown in Figure \ref{fig:intro}, a robust model needs to scan through the video and compare the video context to find the second occurrence of ``child touches the ground". This raises the first challenge for our task: \textit{semantic misalignment}. A simple ordinal number will result in searching from a whole video, where a naive sliding approach will fail. On the other hand, the language query usually describes complex temporal dependencies. Consider another query like "Child is running away after is closest to the camera", different from the sequence described in sentence, the "close to the camera" moment happens before "running away". This raises the second challenge for our task: \textit{structural misalignment}. The language sequence is often misaligned with video sequence, where a naive matching without temporal reasoning will fail. 

    These two key challenges we identify: semantic misalignment and structural misalignment have not been solved in existing methods~\cite{hendricks2017localizing,gao2017tall} for the natural language moment retrieval task. Existing methods sample candidate moments by scanning videos with varying sliding windows, and compare the sentence with each moment individually in a multi-modal common space. Although simple and intuitive, this individualist representations of sentence and video make it hard to model semantic and structural relations among two modalities. 
    
    To address the above challenges, we propose an end-to-end Moment Alignment Network (MAN) for the natural language moment retrieval task. The proposed MAN model directly generates candidate moment representations aligned with language semantics, and explicitly model temporal relationships among different moments in a graph-structured network. Specifically, we encode the entire video stream using a hierarchical convolutional network and naturally assign candidate moments over different temporal locations and scales. Language features are encoded as efficient dynamic filters and convolved with input visual representations to deal with semantic misalignment. In addition, we propose an Iterative Graph Adjustment Network (IGAN) adopted from Graph Convolution Network (GCN)~\cite{kipf2017semi} to model relations among candidate moments in a structured graph. Our contributions are as follows:
    \begin{itemize}
        \vspace{-0.2cm}
        \item We propose a novel single-shot model for the natural language moment retrieval task, where language description is naturally integrated as dynamic filters into an end-to-end trainable fully convolutional network.
        \vspace{-0.2cm}
        \item To the best of our knowledge, we are the first to exploit graph-structured moment relations for temporal reasoning in videos, and we propose the IGAN model to explicitly model temporal structures and improve moment representation.
        \vspace{-0.2cm}
        \item We conduct extensive experiments on two challenging benchmarks: Charades-STA~\cite{gao2017tall} and DiDeMo~\cite{hendricks2017localizing}. We demonstrate the effectiveness of each component and the proposed MAN significantly outperforms the state-of-the-art by a large margin.
    \end{itemize}

\begin{figure*}
\begin{center}
\includegraphics[width=0.95\linewidth]{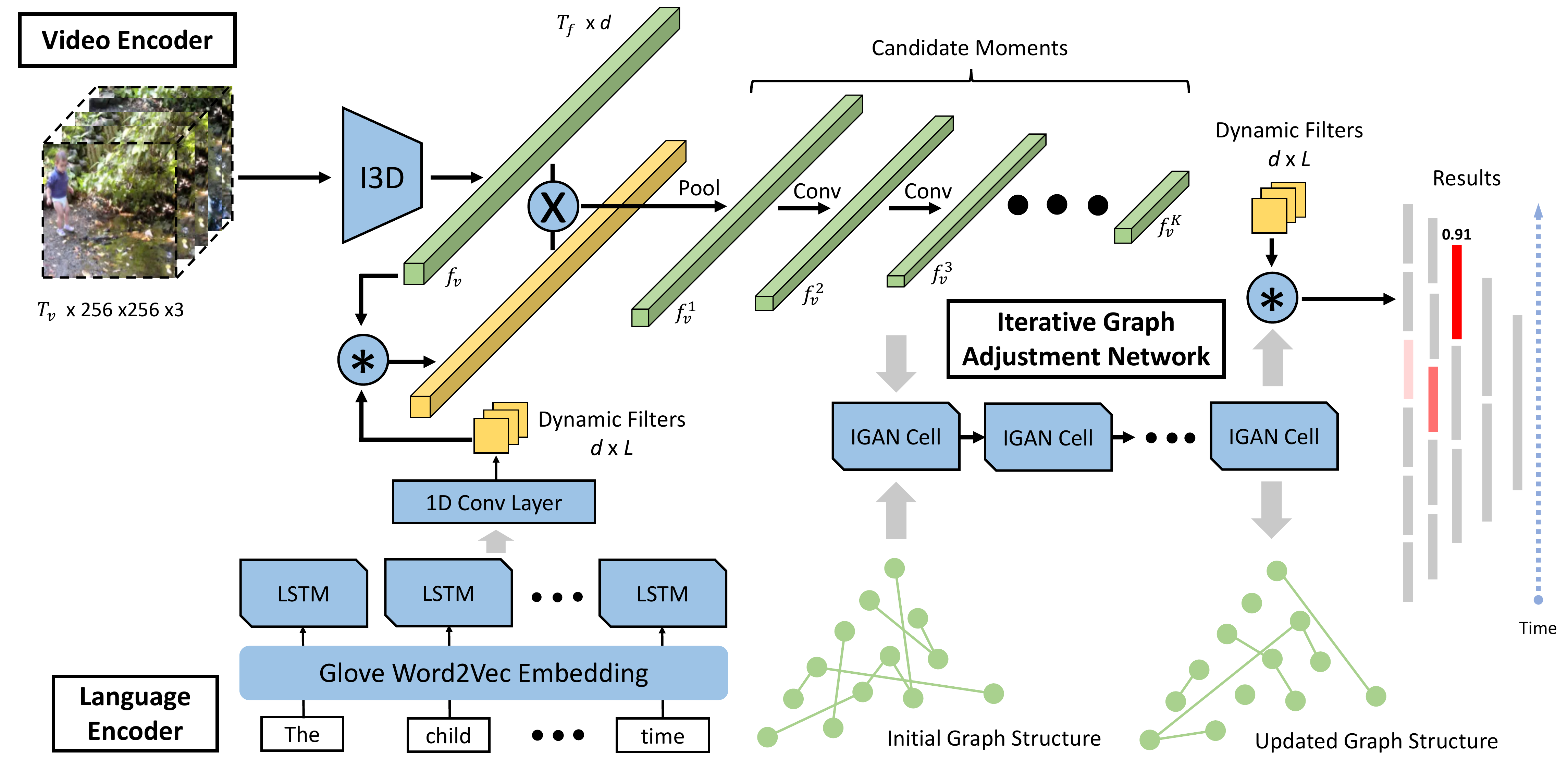}
\end{center}
   \caption{An overview of our end-to-end Moment Alignment Network (MAN) for natural language moment retrieval (best viewed in color). MAN consists three major components: (1) A language encoder to convert the input language query to dynamic convolutional filters through a single-layer LSTM. (2) A video encoder to produce multi-scale candidate moment representations in a hierarchical fully-convolutional network, where input visual features are aligned with language semantics by convolution. (3) An iterative graph adjustment network to directly model moment-wise temporal relations and update moment representations. Finally, the moments are retrieved by its matching scores with the language query.}
\label{fig:overall}
\vspace{-0.4cm}
\end{figure*}  

\section{Related Work}
    \noindent\textbf{Temporal Activity Detection.} Temporal activity detection is the task to predict the start and end time (\ie, localization) and the label (\ie, classification) of activity instances in untrimmed videos. Earlier works on activity detection mainly used temporal sliding windows as candidates and trained activity classifier on hand-crafted features~\cite{oneata2013action,gaidon2013temporal,jain2014action,mettes2015bag,tang2013combining}. With the vast successes of deep learning methods, two-stream networks~\cite{simonyan2014two,feichtenhofer2016convolutional,wang2015towards}, 3D ConvNet~\cite{tran2015learning} and other deep neural networks~\cite{carreira2017quo, tran2018closer, qiu2017learning, NonLocal2018, tan} have been proposed to model video sequences and significantly improved recognition performance. To better localize temporal boundaries, a large body of work incorporated deep networks into the detection framework and obtained improved performance~\cite{chao2018rethinking,lin2017single,dai2017temporal,shou2017cdc,buch2017end,zhao2017temporal,shou2016temporal,xu2017r,zhang2018s3d}. Among these works,  S-CNN~\cite{shou2016temporal} proposed a multi-stage CNN which adopted 3D ConvNet with multi-scale sliding window; R-C3D~\cite{xu2017r} proposed an end-to-end trainable activity detector based on Faster-RCNN~\cite{ren2015faster}; S$^3$D~\cite{zhang2018s3d} performed single-shot activity detection to get rid of explicit temporal proposals.
    
    However, most of these methods have focused on detecting a fixed set of activity classes without language queries. In this paper, we propose to build a highly-integrated retrieval framework and adopt a similar single-shot encoding scheme inspired by the single-shot detectors~\cite{liu2016ssd,zhang2018s3d,lin2017single}. 
    
    \noindent\textbf{Natural Language Moment Retrieval.} The natural language moment retrieval is a new task introduced recently~\cite{hendricks2017localizing,gao2017tall}. The methods proposed in~\cite{hendricks2017localizing,gao2017tall} learn a common embedding space shared by video segment features and sentence representations and measure their similarities through sliding window~\cite{gao2017tall} or handcrafted heuristics~\cite{hendricks2017localizing}. While simple and effective, these methods fail to consider the challenging alignment problems.
    
    Until recently, several methods were proposed to closely integrate language and video representation~\cite{xu2019multilevel,chen2018temporally}: Xu \etal~\cite{xu2019multilevel} proposed multilevel language and video feature fusion; TGN~\cite{chen2018temporally} applied frame-by-word interactions between video and language and obtained improved performance. Although these works share the same spirit with ours to better align semantic information, they fail to reason the complex cross-modal relations. Our work is the first to model both semantic and structural relations together in an unified network, allowing us to directly learn the complex temporal relations in an end-to-end manner.
    
    \noindent\textbf{Visual Relations and Graph Network.} Reasoning about the pairwise relationships has been proven to be very helpful in a variety of computer vision tasks~\cite{gupta2009observing,yao2010modeling,yao2012describing, dai2017fason}. Recently, visual relations have been combined with deep neural networks in areas such as object recognition~\cite{hu2018relation,dollardetecting}, visual question answering~\cite{santoro2017simple} and action recognition~\cite{ma2017attend,ni2016progressively}. A variety of papers have considered modeling spatial relations in natural images~\cite{dai2017detecting,hu2017modeling,peyre2017weakly}, and scene graph is widely used in the image retrieval tasks~\cite{johnson2015image,yang2018graph}. In the field of natural language moment retrieval: Liu \etal~\cite{liu2018temporal} proposed to parse sentence structure as a dependency tree and construct a temporal modular network accordingly; Hendricks \etal~\cite{hendricks2018emnlp} modeled video context as a latent variable to reason about the temporal relationships. However, their reasoning relies on a hand-coded structure, thus, fail to directly learn complex temporal relations.
    
    Our work is inspired by the GCN~\cite{kipf2017semi} and other successful graph-based neural networks~\cite{marino2017more,yan2018spatial}. While the original GCN is proposed to reason on a fixed graph structure, we modify the architecture to jointly optimize relations together. That is, instead of fixing the temporal relations, we learn it from the data.

\section{Model}
    In this work, we address the natural language moment retrieval task. Given a video and a natural language description as a query, we aim to retrieve the best matching temporal segment (\ie, moment) as specified by the query. To specifically handle the semantic and structural misalignment between video and language, we propose Moment Alignment Network (MAN), a novel framework combining both video and language information in a single-shot structure to directly output matching scores between moment and language query through temporal structure reasoning. As illustrated in Figure~\ref{fig:overall}, our model consists of three main components: a language encoder, a video encoder and an iterative graph adjustment network. We introduce the details of each component and network training in this section.
    

\subsection{Language Encoding as Dynamic Filters}
\label{subsec:language}
    Given an input of a natural language sentence as a query that describes the moment of interest, we aim to encode it so that we can effectively retrieve specific moment in video. Instead of encoding each word with a one-hot vector or learning word embeddings from scratch, we rely on word embeddings obtained from a large collection of text documents. Specifically, we use the Glove~\cite{pennington2014glove} word2vec model pre-trained on Wikipedia. It enables us to model complex linguistic relations and handle words beyond the ones in the training set. To capture language structure, we use  a single-layer LSTM network~\cite{hochreiter1997long} to encode input sentences. In addition, we leverage the LSTM outputs at all time steps to seek  more fine-grained interactions between language and video. We also study the effects of using word-level or sentence-level encoding in our ablation study.
    
    In more detail, a language encoder is a function $F_l(\omega)$ that maps a sequence of words $\omega=\{w_i\}_{i=1}^{L}$ to a semantic embedding vector $f_l \in\mathbb{R}^{L\times d}$, where $L$ is the number of words in a sentence and $d$ is the feature dimension, and $F_l$ is parameterized by Glove and LSTM in our case.
    
    Moreover, to transfer textual information to the visual domain, we rely on dynamic convolutional filters as earlier used in ~\cite{li2017tracking,gavrilyuk2018actor}. Unlike static convolutional filters that are used in conventional neural networks, dynamic filters are generated depending on the input, in our case on the encoded sentence representation. As a general convolutional layer, dynamic filters can be easily incorporated with the video encoder as an efficient building block.

    Given a sentence representation $f_l\in\mathbb{R}^{L\times d}$, we generate a set of word-level dynamic filters $\{\Gamma_i\}_{i=1}^L$ with a single fully-connected layer: 
    \begin{equation}
	    \Gamma_i=tanh(W_\Gamma f_l^i + b_\Gamma)
        \label{eq:dynamicf}
    \end{equation}
    where $f_l^i\in\mathbb{R}^{d}$ is the word-level representation at index $i$, and for simplicity, $\Gamma_i$ is designed to have the same number of intput channels as $f_l^i$. Thus, by sharing the same transformation for all words, each sentence representation $f_l\in\mathbb{R}^{L\times d}$ can be converted to a dynamic filter $\Gamma\in\mathbb{R}^{d\times L}$ through a single 1D convolutional layer.
    
    As illustrated in Figure~\ref{fig:overall}, we convolve the dynamic filters with the input video features to produce a semantically-aligned visual representation, and also with the final moment-level features to compute the matching scores. We detail our usage in Section~\ref{subsec:temporal} and Section~\ref{subsec:IGAN}, respectively.
    
\subsection{Single-Shot Video Encoder}
\label{subsec:temporal}
    Existing solutions for natural language moment retrieval heavily relies on handcrafted heuristics~\cite{hendricks2017localizing} or temporal sliding windows~\cite{gao2017tall} to generate candidate segments. However, the temporal sliding windows are typically too dense and often times designed with multiple scales, resulting in a heavy computation cost. Processing each individual moment separately also fails to efficiently leverage semantic and structural relations between video and language.
    
    Inspired by the single-shot object detector~\cite{liu2016ssd} and its successful applications in temporal activity detection~\cite{zhang2018s3d,lin2017single}, we apply a hierarchical convolutional network to directly produce multi-scale candidate moments from the input video stream. Moreover, for the natural language moment retrieval task, the visual features itself undoubtedly play the major role in generating candidate moments, while the language features also help to distinguish the desired moment from others. As such, a novel feature alignment module is especially devised to filter out unrelated visual features from language perspective at an early stage. We do so by generating convolutional dynamic filters (Section~\ref{subsec:language}) from the textual representation and convolving them with the visual representations. Similar to other single shot detectors, all these components are elegantly integrated into one feed-forward CNN, aiming at naturally generating variable-length candidate moments aligned with natural language semantics.
    
    In more detail, given an input video, we first obtain a visual representation that summarizes spatial-temporal patterns from raw input frames into high-level visual semantics. Recently, Dai \etal proposed to decompose 3D convolutions into aggregation blocks to better exploit the spatial-temporal nature of video. We adopt the TAN~\cite{tan} model to obtain a visual representation from video. As illustrated in Figure~\ref{fig:overall}, an input video $V=\{v_t\}_{t=1}^{T_v}$ is encoded into a clip-level feature ${f_v}\in\mathbb{R}^{T_f\times d}$ where $T_f$ is the total number of clips and $d$ is the feature dimension. For simplicity, we set $f_v$ and $f_l$ to have the same number of channels. While ${f_v}$ should be sufficient for building advanced recognition model~\cite{xu2017r,lin2017single,zhang2018dynamic}, the crucial alignment information between language and vision is missing specifically for natural language moment retrieval. 
    
    As such, the dynamic convolutional filters are applied to fill the gap.
    We convolve the dynamic filter $\Gamma$ with ${f_v}$ to obtain a clip-wise response map $M$, and $M$ is further normalized to augment the visual feature. Formally, the augmented feature $f_v'$ is computed as:
    \begin{equation}
    \begin{split}
	    & M = \Gamma * f_v \in \mathbb{R}^{T_v\times L} \\
	    & M_{norm} = softmax(sum(M)) \in \mathbb{R}^{T_v} \\
	    & f_v' = M_{norm} \odot f_v \in \mathbb{R}^{T_v\times d}
	\end{split}
	\label{eq:select}
    \end{equation}
    where $\odot$ denotes matrix-vector multiplication.
    
    To generate variable-length candidate moments, we follow similar design of other single-shot detectors~\cite{liu2016ssd,zhang2018s3d} to build a multi-scale feature hierarchy. Specifically, a temporal pooling layer is firstly devised on top of $f_v'$ to reduce the temporal dimension of feature map and increase temporal receptive field, producing the output feature map of size ${T_v}/p\times d$ where $p$ is the pooling stride. Then, we stack $K$ more 1D convolutional layers (with appropriate pooling) to generate a sequence of feature maps that progressively decrease in temporal dimension which we denote as $\{f_v^k\}_{k=1}^{K}, f_v^k \in\mathbb{R}^{T_k\times d}$ where $T_k$ is the temporal dimension of each layer. Thus each temporal feature cell is responsive to a particular location and length, and therefore corresponds to a specific candidate moment. 

\subsection{Iterative Graph Adjustment Network}
\label{subsec:IGAN}
    To encode complex temporal dependencies, we propose to model moment-wise temporal relations in a graph to explicitly utilize the rich relational information among moments. Specifically, candidate moments are represented by nodes, and their relations are defined as edges. Since we gather $N=\sum_{k=1}^{K}{T_k}$ candidate moments in total each represented by a d-dimensional vector, we denote the node feature matrix as  $f_m \in\mathbb{R}^{N\times d}$. To perform reasoning on the graph, we aim to apply the GCN proposed in~\cite{kipf2017semi}. Different from the standard convolutions which operate on a local regular grid, the graph convolutions allow us to compute the response of a node based on its neighbors defined by the graph relations. In the general form, one layer of graph convolutions is defined as: 
    \begin{equation}
	    H=ReLU(GXW)
        \label{eq:gcn}
    \end{equation}
    where $G\in\mathbb{R}^{N\times N}$ is the adjacency matrix, $X\in\mathbb{R}^{N\times d}$ is the input features of all nodes, $W\in\mathbb{R}^{d\times d}$ is the weight matrix and $H\in\mathbb{R}^{N\times d}$ is the updated node representation. 
    
    However, one major limitation of the GCN applied in our scenario is that it can only reason on a fixed graph structure. To fix this issue, we introduce the Iterative Graph Adjustment Network (IGAN), a framework based on GCN but with a learnable adjacency matrix, that is able to simultaneously infer a graph by learning the weight of all edges and update each node representation accordingly. In more detail, we iteratively updates the adjacency matrix as well as node features in a recurrent manner. The IGAN model is fully differentiable thus can be efficiently learned from data in an end-to-end manner.
    
\begin{figure}
\begin{center}
\includegraphics[width=0.95\linewidth]{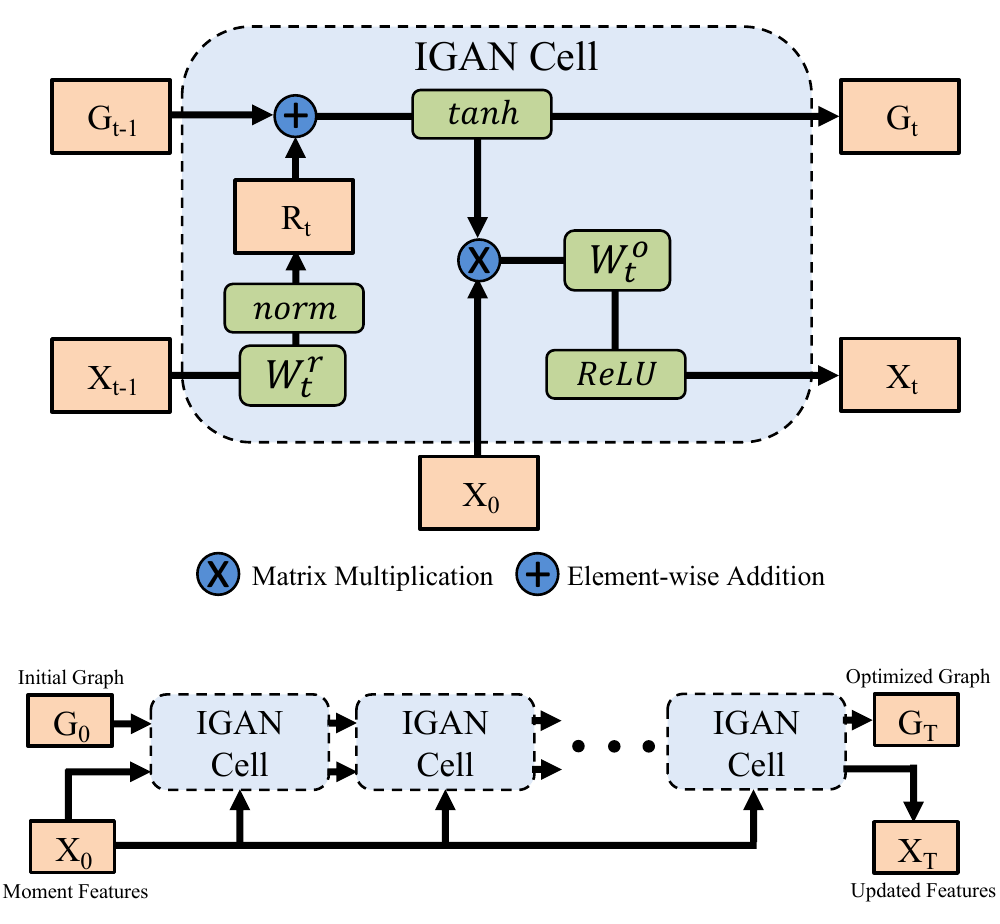}
\end{center}
   \caption{The structure of the proposed Iterative Graph Adjustment Network (IGAN). Top: In each IGAN cell, a residual component $R_t$ is generated from the previous node representation $X_{t-1}$ and aggregated with the preserving component $G_{t-1}$ to produce the current adjacency matrix $G_t$. Node representations are updated according to Equation~\ref{eq:gcn} with $G_t$, $X_0$ and $W_t^o$. Bottom: Multiple IGAN cells are connected to simultaneously update node representation and graph structure.} 
\label{fig:IGAN}
\end{figure}

    In order to jointly learn the node representation and graph structure together, we propose certain major modifications to the original GCN block: (1) Inspired by the successful residual network~\cite{he2016deep}, we decompose the adjacency matrix into a preserving component and a residual component. (2) The residual component is produced from the node representation similar to a decomposed correlation~\cite{carreira2012semantic}. (3) In a recurrent manner, we iteratively accumulate residual signals to update the adjacency matrix by feeding updated node representations. The overall architecture of a single IGAN cell is illustrated in the top half of Figure~\ref{fig:IGAN} and the transition function is formally defined as:
    \begin{equation}
    \begin{split}
	    & R_t = norm(X_{t-1} W_t^r X_{t-1}^T) \\
	    & G_t = tanh(G_{t-1} + R_t) \\
	    & X_t = ReLU(G_t X_0 W_t^o) \\
	\end{split}
	\label{eq:IGAN}
    \end{equation}
    where $X_0=f_m$ is the input candidate moment features, $R_t$ is the residual component derived from the output of previous cell $X_{t-1}$, $norm()$ denotes a signed square root followed by a L2 normalization to normalize the features, and $W_t^r$ and $W_t^o$ are learnable weights. Note that the candidate moment features $X_0$ is the output of a hierarchical convolutional network combined with language information, thus can be jointly updated with the IGAN. 

    In our design, the initial adjacency matrix $G_0$ is set as a diagonal matrix to emphasize self-relations.  we stack multiple IGAN cells as shown in the bottom half of Figure~\ref{fig:IGAN} to update the candidate moment representations as well as the moment-wise graph structure. Finally, we convolve the dynamic filter $\Gamma$ with the final output $X_T$ to compute the matching scores. We further study the effects of IGAN in our ablation study.
    
\subsection{Training}
\label{subsec:training}
    Our training sample consists of an input video, an input language query and a ground truth best matching moment annotated with start and end time. During training, we need to determine which candidate moments correspond to a ground truth moment and train the network accordingly. Specifically, for each candidate moment, we compute the temporal IoU score with ground truth moment. If the temporal IoU is higher than $0.5$, we regard the candidate moment as positive, otherwise negative. After matching each candidate moment with the ground truth, we derive a ground truth matching score $s_i$ for each candidate moment.
     
    For each training sample, the network is trained end-to-end with a binary classification loss using sigmoid cross-entropy. Rather than using a hard score, we use the temporal IoU score $s_i$ as ground truth for each candidate moment. The loss is defined as:
    \begin{equation}
	    \mathcal{L} = -\frac{1}{N_b}\sum_{i}^{N_b}(s_{i}\log({a}_{i}) + (1-s_{i})\log(1-{a}_{i}))
        \label{eq:loss}
    \end{equation}
    where $N_b$ is the number of total training candidate moments in a batch, $a_i$ is the predicted score and $s_i$ is the ground truth score. 

\section{Experiments}
    We evaluate the proposed approach on two recent large-scale datasets for the natural language moment retrieval task: DiDeMo~\cite{hendricks2017localizing} and Charades-STA~\cite{gao2017tall}. In this section we first introduce these datasets and our implementation details and then compare the performance of MAN with other state-of-the-art approaches. Finally, we investigate the impact of different components via a set of ablation studies and provide visualization examples.
    
\subsection{Datasets}
    \noindent\textbf{DiDeMo} The DiDeMo dataset was recently proposed in~\cite{hendricks2017localizing}, specially for natural language moment retrieval in open-world videos. DiDeMo contains more than $10,000$ videos with $33,005$, $4,180$ and $4,021$ annotated moment-query pairs in the training, validation and testing datasets respectively. To annotate moment-query pairs, videos in DiDeMo are trimmed to a maximum of $30$ seconds, divided into $6$ segments of $5$ seconds long each, and each moment contains one or more consecutive segments. Therefore, there are $21$ candidate moments in each video and the task is to select the moment that best matches the query.
    
    
    Following~\cite{hendricks2017localizing}, we use Rank-1 accuracy (Rank@1), Rank-5 accuracy (Rank@5) and mean Intersection-over-Union (mIoU) as our evaluation metrics. 
    
    \noindent\textbf{Charades-STA} The Charades-STA~\cite{gao2017tall} was another recently collected dataset for natural language moment retrieval in indoor videos. Charades-STA is built upon the original Charades~\cite{sigurdsson2016hollywood} dataset. While Charades only provides video-level paragraph description, Charades-STA applies sentence decomposition and keyword matching to generate moment-query annotation: language query with start and end time. Each moment-query pair is further verified by human annotators. In total, there are $12,408$ and $3,720$ moment-query pairs in the training and testing datasets respectively. Since there is no pre-segmented moments, the task is to localize a moment with predicted start and end time that best matches the query.
    
    We follow the evaluation setup in~\cite{gao2017tall} to compute "R@$n$, IoU@$m$", defined as the percentage of language queries having at least one correct retrieval (temporal IoU with ground truth moment is larger than $m$) in the top-n retrieved moments. Following standard practice, we use $n\in\{1, 5\}$ and $m\in\{0.5, 0.7\}$.
    
\subsection{Implementation Details}
    We train the whole MAN model in an end-to-end manner, with raw video frames and natural language query as input. For \textit{language encoder}, each word is encoded as a 300-dimensional Glove word2vec embedding. All the word embeddings are fixed without fine-tuning and each sentence is truncated to have a maximum length of $15$ words. A single-layer LSTM with $d=512$ hidden units is applied to obtain the sentence representation. For \textit{video encoder}, TAN~\cite{tan} is used for feature extraction. The model takes as input a clip of $8$ RGB frames with spatial size $256\times 256$ and extracts a 2048-dimensional representation as output of an average pooling layer. We add another 1D convolutional layer to reduce the feature dimension to $d=512$. Each video is decoded at $30$ FPS and clips are uniformly sampled among the whole video. On Charades, we sample $T_f=256$ clips and set the pooling stride $p=16$ and apply a sequence of 1D convolutional filters (pooling stride $2$) to produce a set of $\{16,8,4,2,1\}$ candidate moments, resulting in $31$ candidate moments in total. Similarly, on DiDeMo, in order to match with the pre-defined temporal boundary, we sample $T_f=240$ clips and set pooling stride $p=40$ with a sequence of 1D convolutional filters (pooling is adjusted accordingly) to produce a set of $\{6,5,4,3,2,1\}$ candidate moments, resulting in $21$ candidate moments in total. For both datasets, we apply $3$ IGAN cells. We implement our MAN on TensorFlow~\cite{abadi2016tensorflow}. The whole system is trained by Adam~\cite{kingma2015adam} optimizer with learning rate $0.0001$.
    
    \begin{table}
\begin{center}
\begin{tabular}{K{2.5cm}|K{1.2cm} K{1.2cm} K{1.2cm}}
\hline
Method & Rank@1 & Rank@5 & mIoU \\
\hline
TMN~\cite{liu2018temporal} & 18.71 & 72.97 & 30.14 \\
TGN~\cite{chen2018temporally} & 24.28 & 71.43 & 38.62 \\
MCN~\cite{hendricks2017localizing} & 24.42 & 75.40 & 37.39 \\
\hline \hline
\textbf{MAN(ours)} & \textbf{27.02} & \textbf{81.70} & \textbf{41.16} \\
\hline
\end{tabular}
\end{center}
\caption{Natural language moment retrieval results on DiDeMo dataset. MAN outperforms previous state-of-the-art mehtods by $\sim3\%$ among all metrics.}
\label{tb:DiDeMo_SOTA}
\end{table}

\begin{table}
\begin{center}
\begin{tabular}{K{1.9cm}|K{1.1cm} K{1.1cm} K{1.1cm} K{1.1cm}}
\hline
Method & R@1 IoU=0.5 & R@1 IoU=0.7 & R@5 IoU=0.5 & R@5 IoU=0.7 \\
\hline
Random~\cite{gao2017tall} & 8.51 & 3.03 & 37.12 & 14.06 \\
CTRL~\cite{gao2017tall} & 21.42 & 7.15 & 59.11 & 26.91 \\
Xu \etal ~\cite{xu2019multilevel} & 35.60 & 15.80 & 79.40 & 45.40 \\
\hline \hline
\textbf{MAN(ours)} & \textbf{46.53} & \textbf{22.72} & \textbf{86.23} & \textbf{53.72} \\
\hline
\end{tabular}
\end{center}
\caption{Natural language moment retrieval results on Charades-STA dataset. MAN significantly outperforms previous state-of-the-art methods by a large margin.}
\label{tb:Charades_SOTA}
\vspace{-0.3cm}
\end{table}

\subsection{Comparison with State-of-the-art}
    We compare our MAN with other state-of-the-art methods on DiDeMo~\cite{hendricks2017localizing} and Charades-STA~\cite{gao2017tall}. Note that the video content and language queries differ a lot among two different datasets. Hence, strong adaptivity is required to perform consistently well on both datasets. Since our MAN only takes raw RGB frames as input and doesn't rely on external motion features such as optical flow, for a fair comparison, all compared methods use RGB features only.
    
    \noindent\textbf{DiDeMo} Table~\ref{tb:DiDeMo_SOTA} shows our natural language moment retrieval results on the DiDeMo dataset. We compare with state-of-the-art methods published recently including the methods that use temporal modular network~\cite{liu2018temporal}, fine-grained frame-by-word attentions~\cite{chen2018temporally} and temporal contextual encoding~\cite{hendricks2017localizing}. Among all three evaluation metrics, the proposed method outperforms previous state-of-the-art methods by around $3\%$ in absolute values.

    \noindent\textbf{Charades-STA} We also compare our method with the recent state-of-the-art methods on Charades-STA dataset. The results are shown in Table~\ref{tb:Charades_SOTA}, where CTRL~\cite{gao2017tall} applies a cross-modal regression localizer to adjust temporal boundaries and Xu \etal~\cite{xu2019multilevel} even boosts the performance with more closely multilevel language and vision integration. Our model tops all the methods among all evaluation metrics and significantly improves R@1, IoU=0.5 by over $10\%$ in absolute values. 
    

\subsection{Ablation Studies}
    To understand the proposed MAN better, we evaluate our network with different variants to study their effects. 

\begin{table}
\begin{center}
\begin{tabular}{K{2.8cm}|K{1.2cm} K{1.2cm} K{1.2cm}}
\hline
Method & Rank@1 & Rank@5 & mIoU \\
\hline
Base & 23.56 & 77.66 & 36.36 \\
\hline
Base+FA(1) & 24.45 & 78.69 & 37.72 \\
Base+FA(L) & 25.10 & 79.57 & 38.78 \\
\hline
Base+FA+IGANx1 & 25.67 & 79.36 & 39.13 \\
Base+FA+IGANx2 & 26.10 & 80.08 & 40.21 \\
Base+FA+IGANx3 & \textbf{27.02} & \textbf{81.70} & \textbf{41.16} \\
\hline
\end{tabular}
\end{center}
\caption{Ablation study for effectiveness of MAN components: Top: Advantage of a single-shot video encoder. Mid: Effectiveness of the feature alignment. Bottom: Importance of the IGAN.}
\label{tb:ablation-comp}
\vspace{-0.3cm}
\end{table}

    \noindent\textbf{Network Components.} On DiDeMo dataset, we perform ablation studies to investigate the effect of each individual component we proposed in this paper: single-shot video encoder, feature alignment with language query and iterative graph adjustment network.

    \textit{Single-shot video encoder.} In this work, we introduced a single-shot video encoder using hierarchical convolutional network for the natural language moment retrieval task. To study the effect of this architecture alone, we build a \textbf{Base} model which is the same as we described in Section~\ref{subsec:temporal} except for two modifications: (1) We remove the feature alignment component (Equation~\ref{eq:select}) and directly use the visual feature $f_v$ to construct the network. (2) We remove all IGAN cells on top and directly feed $f_m$ to compute matching scores. The result is reported in the top line in Table~\ref{tb:ablation-comp}, even with only a single-shot encoding scheme, we achieve $23.56\%$ on Rank@1 and $77.66\%$ on Rank@5 which is better or competitive with other state-of-the-art methods. 

\begin{table}
\begin{center}
\begin{tabular}{K{1.9cm}|K{1.1cm} K{1.1cm} K{1.1cm} K{1.1cm}}
\hline
Method & R@1 IoU=0.5 & R@1 IoU=0.7 & R@5 IoU=0.5 & R@5 IoU=0.7 \\
\hline
Xu \etal ~\cite{xu2019multilevel} & 35.60 & 15.80 & 79.40 & 45.40 \\
MAN-VGG & 41.24 & 20.54 & 83.21 & 51.85 \\
MAN-TAN & \textbf{46.53} & \textbf{22.72} & \textbf{86.23} & \textbf{53.72} \\
\hline
\end{tabular}
\end{center}
\caption{Ablation study on different visual features. MAN with VGG-16 features already outperforms state-of-the-art method, and TAN features further boost the performance.}
\label{tb:ablation-visual}
\vspace{-0.3cm}
\end{table}

    \textit{Dynamic filter.} We further validate our design to augment the input clip-level features with dynamic filters. The results are shown in the middle part in Table~\ref{tb:ablation-comp}. On top of the Base model, we study two different variants: (1) Construct a sentence-level dynamic filter where only the last LSTM hidden state is used for feature alignment, denoted as \textbf{Base+FA(1)}. (2) Construct word-level dynamic filters where all LSTM hidden states are converted to a multi-channel filter for feature alignment, denoted as \textbf{Base+FA(L)}. We observe that Base+FA(1) already improves the accuracy compared to the base model, which indicates the importance of adding feature alignment in our model. Moreover, adding more fine-grained word-level interactions between video and language can further improve the performance.

\begin{figure*}
\begin{center}
\includegraphics[width=0.95\linewidth]{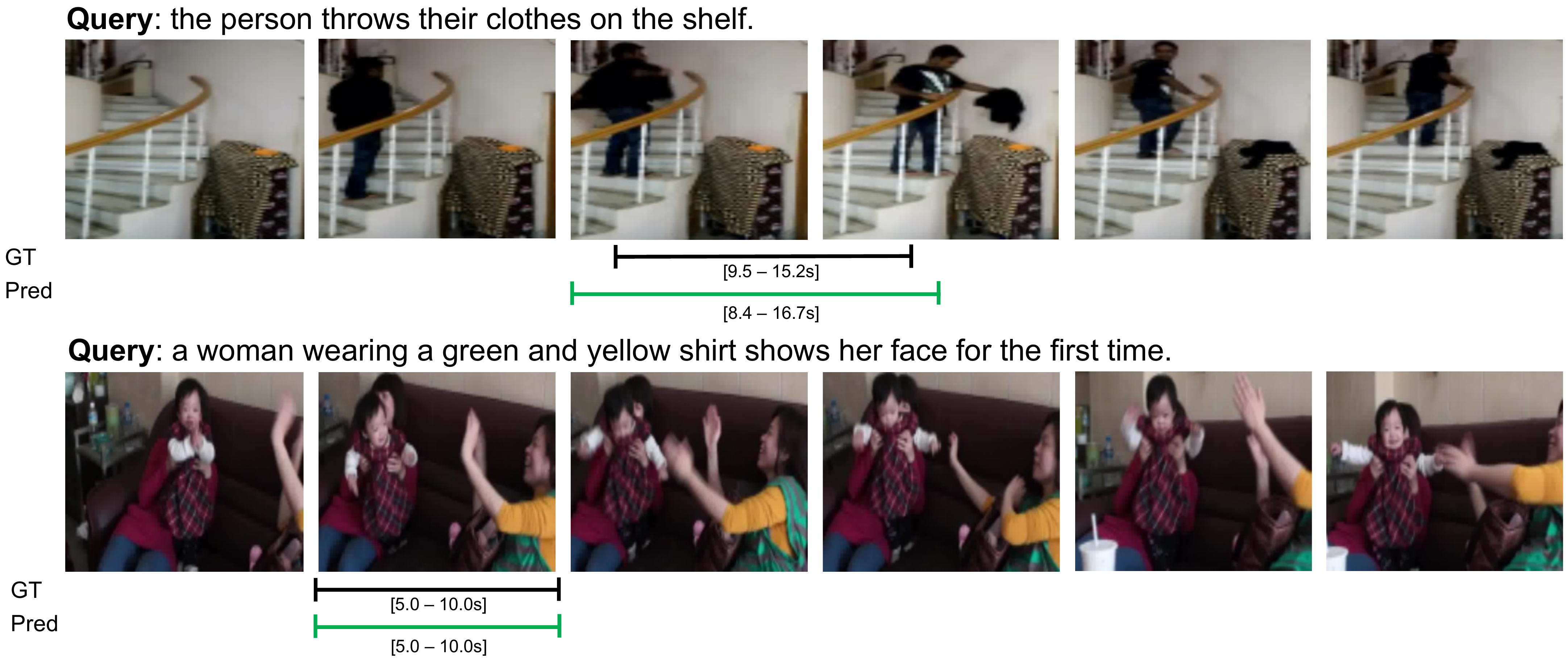}
\end{center}
   \caption{Qualitative visualization of the natural language moment retrieval results (Rank@1) by MAN (best viewed in color). First example is from Charade-STA dataset, and second example is from DiDeMo dataset. Ground truth moments are marked in black and retrieved moments are marked in green.} 
\label{fig:qualitative}
\vspace{-0.4cm}
\end{figure*}   

    \textit{Iterative graph adjustment network.} A major contribution of MAN is using the IGAN cell to iteratively update graph structure and learned representation. We measure the contribution of this component to the retrieval performance in the bottom section in Table~\ref{tb:ablation-comp}, where \textbf{Base+FA+IGANxn} denotes our full model with $n$ IGAN cells. The result shows a decrease in performance with fewer IGAN cells, dropping monotonically from $27.02\%$ to $25.67\%$ on Rank@1. This is because the temporal relations represented in a moment graph structure can be iteratively optimized thus more IGAN cells result in better representation for each candidate moment. Despite the performance gain, we also notice that Base+FA+IGANx3 converges faster and generalizes better with smaller variance.
    
    
\begin{figure}
\begin{center}
\includegraphics[width=0.95\linewidth]{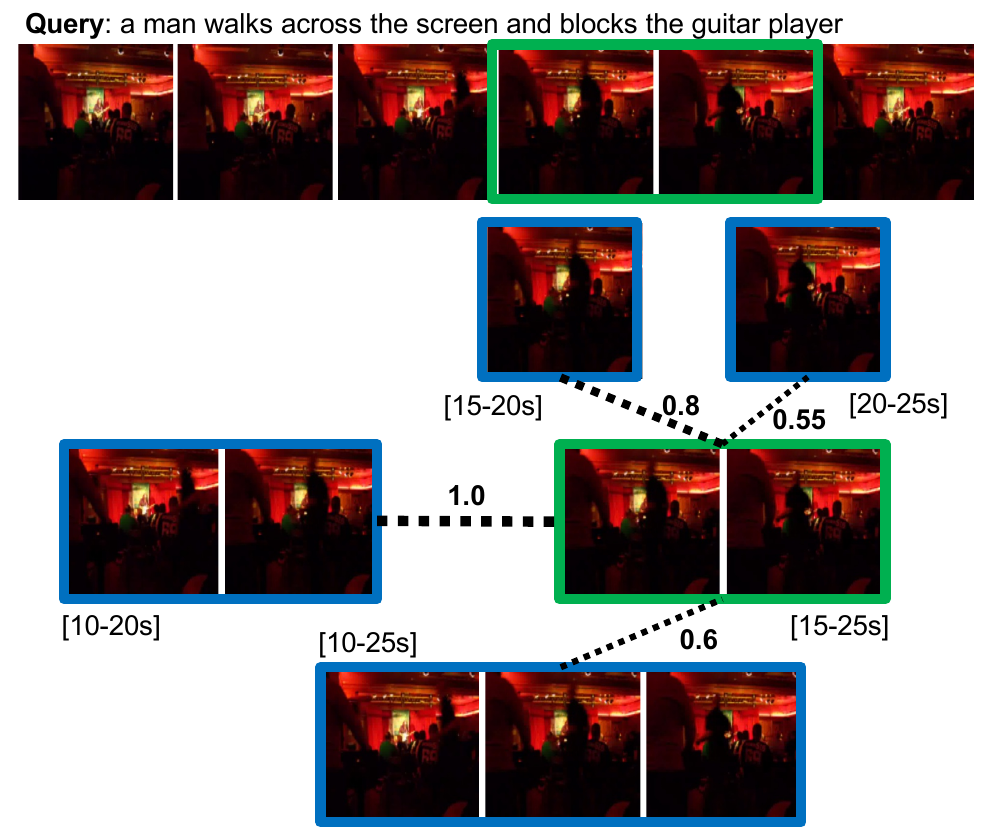}
\end{center}
   \caption{Qualitative example of MAN evaluated on a video-query pair (best viewed in color). The final moment-wise graph structure with top related edges and their corresponding moments is visualized. The retrieved moment is marked in green and other moments are marked in blue. The dashed line indicates the strength of each edge with the highest one normalized to $1.0$.} 
\label{fig:visualize}
\vspace{-0.2cm}
\end{figure}

    \noindent\textbf{Visual Features.} We conduct experiments to study the effect of different visual features on Charades-STA dataset. We consider two different visual features: (1) Two-stream RGB features~\cite{simonyan2014two} from the original Charades dataset, which is a frame-level feature from VGG-16~\cite{simonyan2014very} network, we denote the model as \textbf{MAN-VGG}. (2) TAN features as described in the paper, which is a clip-level feature from aggregation blocks, we denote the model as \textbf{MAN-TAN}. The results are summarized in Table~\ref{tb:ablation-visual}. It can be seen that TAN features outperform VGG-16 features among all evaluation metrics, this is consistent with the fact that better base network leads to better overall performance. But more interestingly, while the overall performance using only VGG visual features is noticeably lower than using TAN features, our \textbf{MAN-VGG} model already significantly outperforms the state-of-the-art method. Since frame-level VGG-16 network provides no motion information when extracting features, this superity highlights MAN's strong ability to perform semantic alignment and temporal structure reasoning. 
    
    \noindent\textbf{Visualization.} \textit{Qualitative Results.} We provide qualitative results to demonstrate the effectiveness and robustness of the proposed MAN framework. As shown in Figure~\ref{fig:qualitative}, MAN is capable of retrieving a diverse set of moments including the one requiring strong temporal dependencies to identify "woman shows her face for the first time". The advantage of MAN is best pronounced for tasks that rely on reasoning complex temporal relations.
    
    \textit{Graph Visualization.} An advantage of a graph structure is its interpretability. Figure~\ref{fig:visualize} visualizes the final moment-wise graph structure learned by our model. In more detail, Figure~\ref{fig:visualize} displays a 30-second video where "man walks" from $10$ to $30$ seconds and "blocks the guitar player" from $15$ to $25$ seconds. MAN is able to concentrate on those moments with visual information related to "man walks across the screen". It also reasons among multiple similar moments including some incomplete moments ($15$-$20$s, $20$-$25$s) and some other moments partially related to "blocks the guitar player" ($10$-$20$s, $10$-$25$s) to retrieve the one best matching result ($15$-$25$s).

\section{Conclusion}
    We have presented MAN, a Moment Alignment Network that unifies candidate moment encoding and temporal structural reasoning in a single-shot structure for natural language moment retrieval. Particularly, we identify two key challenges (\ie semantic misalignment and structural misalignment) and study how to handle such challenges in a deep learning framework. To verify our claim, we propose a fully convolutional network to force cross-modal alignments and an iterative graph adjustment network is devised to model moment-wise temporal relations in an end-to-end manner. With this framework, We achieved state-of-the-art performance on two challenging benchmarks Charades-STA and DiDeMo. 
{\small
\bibliographystyle{ieee_fullname}
\bibliography{egbib}
}

\end{document}